\documentclass[letterpaper]{article} 
\usepackage{aaai25}  
\usepackage{times}  
\usepackage{helvet}  
\usepackage{courier}  
\usepackage[hyphens]{url}  
\usepackage{graphicx} 
\usepackage{natbib}  
\usepackage{caption} 
\usepackage{algorithm}
\usepackage{algorithmic}

\usepackage{newfloat}
\usepackage{listings}
\DeclareCaptionStyle{ruled}{labelfont=normalfont,labelsep=colon,strut=off} 
\floatstyle{ruled}
\newfloat{listing}{tb}{lst}{}
\floatname{listing}{Listing}
\title{Reducing Hallucinations of Medical Multimodal Large Language Models with Visual Retrieval-Augmented Generation}

\author {
    Yun-Wei Chu\textsuperscript{\rm 1,2}\equalcontrib,
    Kai Zhang\textsuperscript{\rm 1,3}\equalcontrib,
    Christopher Malon\textsuperscript{\rm 1},
    Martin Renqiang Min\textsuperscript{\rm 1}
}
\affiliations {
    \textsuperscript{\rm 1}NEC Laboratories America, \textsuperscript{\rm 2}Purdue University, \textsuperscript{\rm 3}Lehigh University\\
    chu198@purdue.edu, kaz321@lehigh.edu, malon@nec-labs.com, renqiang@nec-labs.com
}

\usepackage{bibentry}

\usepackage{amsmath}
\usepackage{amssymb}
\usepackage{graphicx}
\usepackage{xspace}
\usepackage{xcolor}
\usepackage{multirow}
\usepackage{bigstrut}
\usepackage{booktabs}
\usepackage{bm}
\usepackage{amssymb}
\usepackage{bbding}

\begin{document}
\maketitle
\begin{abstract}

Multimodal Large Language Models (MLLMs) have shown impressive performance in
vision and text tasks.  However, hallucination remains a major challenge,
especially in fields like healthcare where details are critical.
In this work, we show how MLLMs may be enhanced to support Visual RAG
(V-RAG), a retrieval-augmented generation framework that incorporates
both text and visual data from retrieved images.  On the MIMIC-CXR chest X-ray
report generation and Multicare medical image caption generation datasets,
we show that Visual RAG improves the accuracy of entity probing,
which asks whether a medical entities is grounded by an image.
We show that the improvements extend both to frequent and rare entities, the latter of which may have less positive training data.
Downstream, we apply V-RAG with entity probing to correct hallucinations and
generate more clinically accurate X-ray reports, obtaining a higher
RadGraph-F1 score.

\end{abstract}

\section{Introduction}
\begin{figure}[t]
    \centering
    \setlength{\abovecaptionskip}{1mm}
    \includegraphics[width=0.9\linewidth]{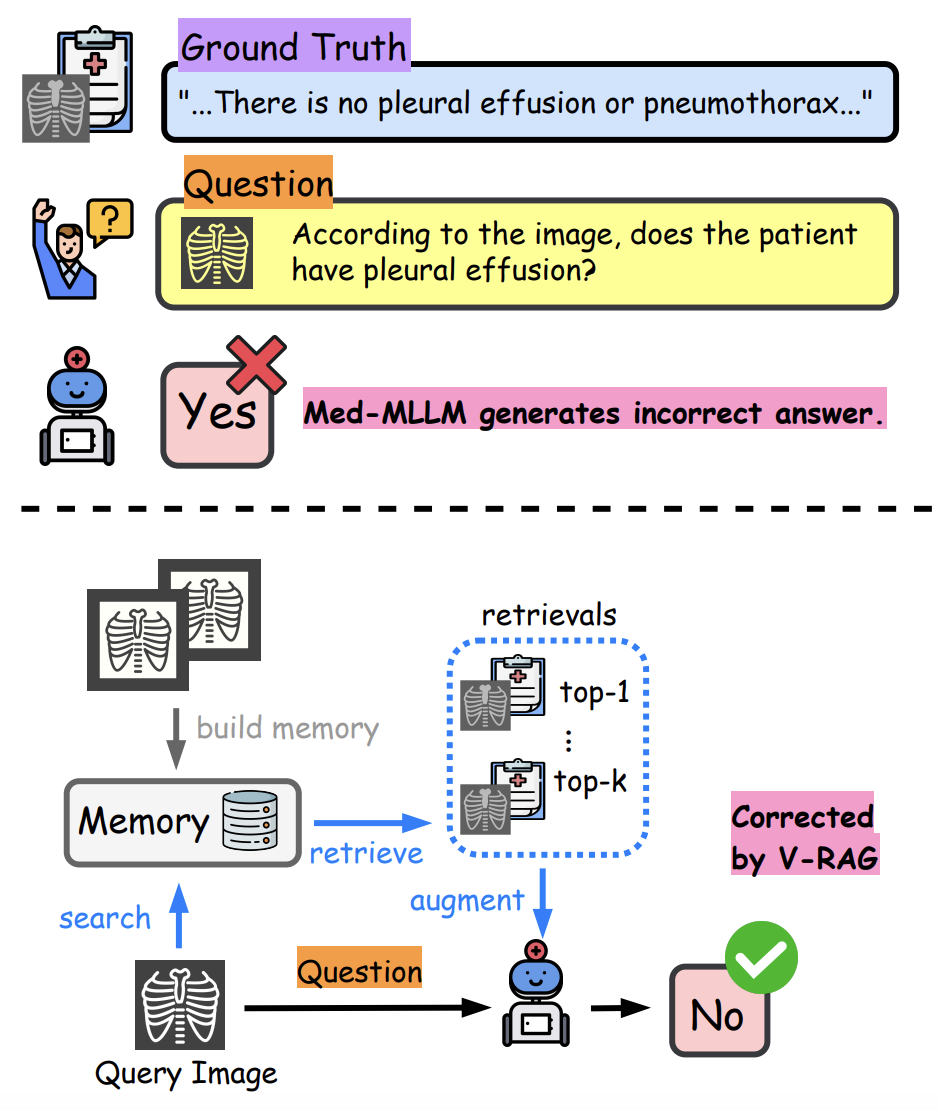}
    \caption{(Up) Hallucination issue of Med-MLLM. (Down) Framework of V-RAG to improve Med-MLLM.}
    \vspace{-5mm}
    \label{fig:fig1}
\end{figure}

Recent advances in Multimodal Large Language Models (MLLMs)~\cite{Achiam2023GPT4TR, Liu2023ImprovedBW} have demonstrated impressive capabilities in complex vision-and-text tasks, showing significant potential in specialized domains.
In healthcare, the development of Medical MLLMs (Med-MLLMs)~\cite{Li2023LLaVAMedTA, Wu2023TowardsGF} can support clinical decision-making processes, with the potential to enhance physician efficiency and improve patient health outcomes.
However, numerous studies have demonstrated that MLLMs are prone to hallucination~\cite{Li2023EvaluatingOH, Bai2024HallucinationOM, Huang2024VisualHO}.
The hallucination tendency of MLLM's has been demonstrated on Med-MLLM's
as well \citep{wu2024hallucinationbenchmarkmedicalvisual}.
This is particularly concerning in the healthcare scenario, as depicted in Figure~\ref{fig:fig1}, where even a few wrong tokens in text can lead to significant misinterpretations, affecting medical diagnoses, treatment plans, and patient outcomes~\cite{Pal2024GeminiGT}.

\begin{figure}[t]
\centering
    \setlength{\abovecaptionskip}{1mm}
    \includegraphics[width=0.9\linewidth]{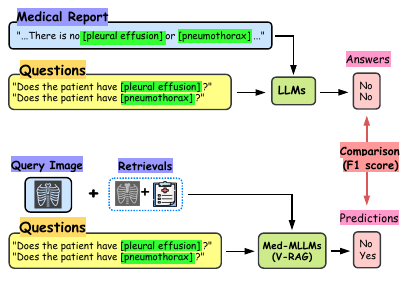}
    \caption{Entity probing asks entity-based questions to an MLLM and compares predictions against answers grounded in an LLM's interpretation of a reference caption.}
    \vspace{-5mm}
    \label{fig:entityprobing}
\end{figure}

Retrieval-Augmented Generation (RAG)~\cite{Lewis2020RetrievalAugmentedGF} has become a prominent approach to mitigate the hallucination problem in Large Language Models (LLMs) by grounding text generation in retrieved knowledge relevant to a given query. 
Besides grounding, RAG potentially supplements the knowledge in a model's parameters with knowledge present in a corpus, enabling open book question answering to exceed closed book performance.
Several prior works~\cite{Sarto2024TowardsRA,Liu2024RARRA, Zhou2024Img2LocRI} have explored text-based RAG in MLLMs.
This approach assumes that using text documents associated with images similar to the query image can effectively augment the model, treating the retrieved images as perfectly interchangeable with the query image.
However, this assumption is not always accurate. In this work, we study \textbf{Visual-RAG (V-RAG)}, which considers not only the associated text from retrieved similar images but also the similar images themselves to provide more accurate responses to the given instruction. By incorporating both modalities, V-RAG allows the model to determine what is truly important from the retrieved content, enhancing its ability to deliver more contextually relevant answers, as illustrated in Figure~\ref{fig:fig1}.

With certain multi-image-trained Med-MLLMs, we see that V-RAG improves a detailed understanding of an image beyond what is possible with text-based RAG techniques.
We demonstrate this through \textbf{entity probing}.
Entity probing presents an image to an MLLM and asks yes/no questions about disease entities, and compares predictions against answers grounded in an LLM's interpretation of a reference report or caption (Figure~\ref{fig:entityprobing}).
Entity probing gives us a clinical perspective on text generations across medical domains which is not captured by natural language generation metrics such as ROUGE, while avoiding sensitivity to entity phrasing.
We show that V-RAG, as an inference technique applied to carefully selected Med-MLLMs trained on multi-image datasets, enhances understanding more effectively than original Med-MLLMs and previous text-based RAG systems.

To improve the model's multimodal understanding when presented with rich retrievals, we design a general fine-tuning technique to boost Med-MLLM capabilities in V-RAG. 
This approach strengthens image-text comprehension and enables effective learning from similar resources retrieved during multimodal queries.
It benefits not only Med-MLLMs trained on multi-image dataset but also single-image-trained models to leverage multi-image inputs in V-RAG, thereby improving performance.
This frees researchers from relying on specific pre-trained models that may not be aligned with their task in order to use V-RAG, allowing V-RAG to be applied to any model and dataset of interest.  
Our key contributions are summarized as follows: 

\begin{itemize}
\item We analyze hallucinations in MLLMs on chest X-ray report generation and medical image captioning datasets through entity probing, showing that V-RAG mitigates hallucinations more effectively than baseline RAG techniques. These benefits extend to both frequent and rare entities.

\item  To enhance Med-MLLMs' multimodal comprehension with V-RAG, we introduce general image-text fine-tuning tasks to boost model performance and improve their understanding when multimodal retrievals are presented. These tasks enable an MLLM originally trained with single images to become capable of V-RAG using multiple retrieved images.

\item We show that entity probing with V-RAG can be used to revise
chest X-ray reports to contain fewer hallucinations and have better 
detailed accuracy, as measured by RadGraph-F1 score.

\end{itemize}

\section{Related Work}

\paragraph{Medical Multimodal Large Language Models.}
Substantial advancements have been made in adapting MLLMs to medical imaging~\cite{Zhang2023PMCVQAVI, Wu2023TowardsGF, Moor2023MedFlamingoAM,Lee2023LLMCXRIL}. The primary focus has been on training these models for radiology tasks using medical images (like X-rays, MRIs, and CT scans) along with their textual descriptions/reports.
\citet{Li2023LLaVAMedTA} used GPT-4 to generate instruction-following data for fine-tuning, improving MLLMs' conversational ability for open-ended biomedical image inquiries.
\citet{Chen2024CheXagentTA} developed a foundation model for chest X-Ray interpretation with an image-text bridger to align modalities.
However, we found that these medical multimodal foundation models still suffer from hallucinations.
We aim to mitigate this issue in Med-MLLMs through a visual-based Retrieval-Augmented Generation (RAG) approach, enabling these models to generate factually accurate answers.

\paragraph{Retrieval-Augmented Generation (RAG).} 
RAG~\cite{Lewis2020RetrievalAugmentedGF} mitigates hallucination in LLMs by retrieving and integrating domain-specific knowledge from external databases, enhancing text generation with accurate, aligned information and effectively addressing this challenge~\cite{Guu2020REALMRL, Siriwardhana2022ImprovingTD, Shahul2023RAGAsAE}.
Despite RAG's popularity, very few studies have applied RAG to MLLMs. 
Prior studies primarily enhance image captioning by reranking labels of retrieved images~\cite{Liu2024RARRA,Qu2024AlleviatingHI} or directly incorporating texts from these images into prompts to improve generation~\cite{Liu2023LearningCV, Sarto2024TowardsRA, Zhou2024Img2LocRI}. 
In healthcare, researchers have developed domain-specific retrieval pipelines~\cite{Sun2024FactAwareMR} and explored the optimal number of retrievals~\cite{Xia2024RULERM} to ensure the factuality of Med-MLLMs.
All these previous works retrieve similar images based on the query image but consider only the text/label associated with the retrieved images.
Thus these methods assume that the retrieved images are perfectly interchangeable with the query image, which is not always the case. 

A more effective approach might involve comparing the query image with retrieved images and their reports, allowing the model to identify what is truly relevant for generation.  This is the ``V-RAG'' method of our paper.
\citet{august} attempted a similar approach with ``Coarse (I+T),'' though it performed worse than using only associated texts (``Coarse (T)'' in their Table 6), which they noted was likely due to limited multi-image reasoning in the MLLMs they considered. 
We address this by analyzing MLLMs trained for multi-image reasoning, and also by introducing an architecture and fine-tuning method to make single-image-trained MLLMs ``V-RAG-capable,'' enabling them to benefit from this approach.

\begin{figure*}[t]
    \centering
    \setlength{\abovecaptionskip}{1mm}
    \includegraphics[width=\linewidth]{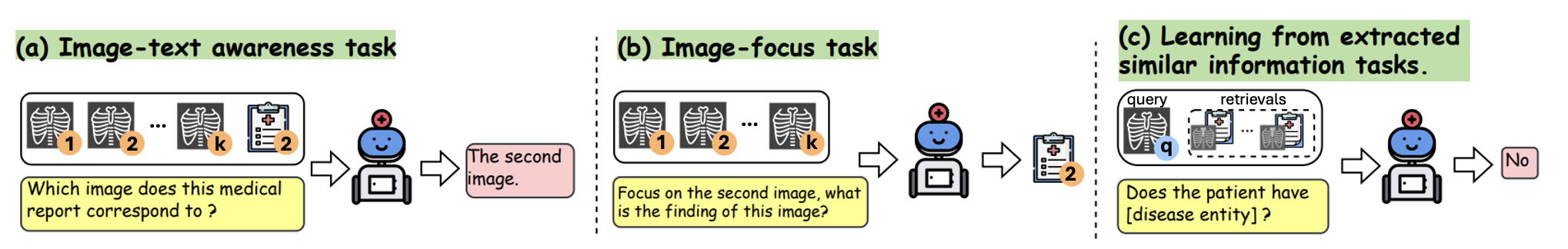}
    \caption{Fine-tuning tasks to make Med-MLLM V-RAG-capable by (a) improving image-and-text association abilities, (b) focusing on specific images, and (c) making decisions using extracted similar data.}
    \vspace{-3mm}
    \label{fig:tasks}
\end{figure*}

\section{V-RAG with Existing multi-image-trained Med-MLLMs}
\label{sec:v-rag-method}
Figure~\ref{fig:fig1} illustrates the V-RAG framework. This section details each component and explains how we enhance model performance during V-RAG.

\subsection{Multimodal Retrieval}

We aim to retrieve images and corresponding textual descriptions that match the features of target medical images. These references, rich in visual and textual medical details, guide response generation for the medical image.
To extract embeddings, we employ BiomedCLIP~\cite{Zhang2023BiomedCLIPAM}, which provides robust representations across a diverse range of biomedical image types. For a given medical image $X_{img}$, we extract its image embedding $\mathcal{E}_{img} \in \mathbb{R}^{d}$, with $d$ representing the dimension (i.e., 512 for BiomedCLIP), and store it in memory $\mathcal{M}$ for retrieval.

To facilitate efficient search operations during the inference phase, we construct the memory $\mathcal{M}$ using FAISS~\cite{douze2024faiss}, a vector storage and retrieval system that utilizes GPU computation. Instead of exact kNN search, we employ an approximate kNN search using the Hierarchical Navigable Small World (HNSW) algorithm~\cite{Malkov2016EfficientAR} to identify the top-$k$ nearest neighbors, effectively retrieving the images in $\mathcal{M}$ most similar to a given query image.

\subsection{Inference with V-RAG}

In the inference stage, we first encode the query image $X_{q}$ to obtain its corresponding image embedding. We then retrieve the top-$k$ images in $\mathcal{M}$; the retrieved set of similar images and their reports are represented as $(I_{1}, ... I_{k})$ and $(R_{1}, ..., R_{k})$.
We then use the retrievals to guide the generation of Med-MLLM for the query image by appending each reference before the question, following this prompt guidance:
``\texttt{...This is the i-th similar image and its report for your reference. [Reference]$_i$... Answer the question with only the word yes or no. Do not provide explanations. According to the last query image and the reference images and reports, [Question] [Query Image]}", where \texttt{[References]$_i$} is structured as \texttt{[($I_i$, $R_i$)]}.

\subsection{Enhancing Med-MLLMs for V-RAG}
Some MLLMs may lack the training to distinguish information from multiple images.
To address this, we introduce three fine-tuning tasks to enhance image-text association in the V-RAG process.
Given a dataset of images paired with captions or reports, we define the original dataset as $S = {(\texttt{img}_{i}, \texttt{P}_i, \texttt{A}_i)}|^{N}_{i=1}$, where $\texttt{img}_{i}$ denotes the $i$-th image, $\texttt{P}_i$ and $\texttt{A}_i$ represent the prompt and the answer, respectively, and $N$ is the total number of samples.
We then construct fine-tuning tasks on this dataset with our designed objectives as follows.

\paragraph{Image-text awareness task.}
We aim to enhance Med-MLLM's image-and-text association ability by training the model to identify the relevant image corresponding to provided text from multiple images.
To achieve this, we construct a multi-image dataset, $M_{position}$, from dataset $S$, to ask the model to identify the position of the image related to the given text, as depicted in Figure~\ref{fig:tasks}(a).
First, we randomly select $K$ images (where $K$ ranges from 1 to 5 in our case) and form the image collection $(\texttt{img}_{i_1}, ..., \texttt{img}_{i_K})$.
Next, we choose an integer $j$ from $[1,K]$ and retrieve the textual document $\texttt{R}_{i_j}$, corresponding to $\texttt{img}_{i_j}$.
We then collect $M_{position}$ using $\{(\texttt{img}_{i_1}, \texttt{img}_{i_2}, ..., \texttt{img}_{i_K}, \texttt{P}^{'}_{i_j}, \texttt{A}^{'}_{i_j})\}$. 
Here, $\texttt{P}^{'}_{i_j}$ is a newly formulated prompt designed to ask a position-based question in addition to the original question $\texttt{P}_{i_j}$, associating $\texttt{A}_{i_j}$ with the provided images.
For example, ``\texttt{What image from 1 to $K$ does this $\texttt{A}_{i_j}$ correspond to? $\texttt{P}_{i_j}$}''.
$\texttt{A}^{'}_{i_j}$ is the answer indicating the position of ${\texttt{img}_{i_j}}$ among the provided images, for example, ``\texttt{The $j$-th image.}''

\paragraph{Image-focus task.}
In this task (Figure~\ref{fig:tasks}(b)), we aim to direct Med-MLLM to focus on one specific image from a set of multiple images and subsequently perform text generation based on that image, thereby improving performance by minimizing distractions from other visual inputs.
To achieve this, we create another dataset, $M_{focus}$, also from image dataset $S$.
We start by randomly selecting $K$ images from $S$ to form the collection $(\texttt{img}_{i_1}, ..., \texttt{img}_{i_K})$, and then choose an integer $j$ from $[1,K]$.
We then collect $\{(\texttt{img}_{i_1}, ..., \texttt{img}_{i_K}, \texttt{P}^{''}_{i_j}, \texttt{A}_{i_j})\}$ to form $M_{focus}$, where  $\texttt{P}^{''}_{i_j}$ is a new prompt designed to help the model focus on our specified image, $\texttt{img}_{i_j}$, and pose the original question $\texttt{P}_{i_j}$ for that image.
For example in Figure~\ref{fig:tasks}(b), the new prompt $\texttt{P}^{''}_{i_j}$ is ``\texttt{Focus on the $j$-th image, {P}$_{i_j}$.}'', where $\texttt{P}_{i_j}$ is the original prompt that asks for a finding/report to be generated from a given image.

\paragraph{Strategies to make easier learning tasks.}
Various conditions may be applied to the random selection of images for both image-text awareness and image-focus tasks.
For example, when the image dataset $S$ consists of images $\texttt{img}_i$ with radiology reports $\texttt{A}_i$, we  require that the selected report $\texttt{A}_{i_j}$ for the focus image contains at least one CheXpert~\cite{Irvin2019CheXpertAL} label that is distinct from those in the other reports $\{A_{{i_m}}\}|^{K}_{m = 1, m \neq j}$.
This strategy simplifies the learning task by ensuring that there are no alternative images to which the report could apply equally well.
For easier and more diverse datasets, such a strategy may not be necessary.

\paragraph{Learning from extracted similar information task.}
We aim to assist Med-MLLM in decision-making by using extracted similar information during V-RAG. 
To do so, we simulate the V-RAG scenario and construct a multi-image dataset, $M_{vrag}$.
Given a query image $\texttt{img}_{q}$ in the validation set, we search for the top-$K$ similar images ($\texttt{img}_{q_1}$, ... , $\texttt{img}_{q_K}$) from memory $\mathcal{M}$, pairing them with their corresponding documents ($\texttt{A}_{q_1}$, ..., $\texttt{A}_{q_K}$).
We then conduct $M_{vrag}$ using $\{(\texttt{img}_{q_1},  \texttt{A}_{q_1}, ..., \texttt{img}_{q_K}, \texttt{A}_{q_K}), \texttt{img}_{q}, \texttt{P}^{'''}_{q}, \texttt{A}_{q}\}$.
Here, $ \texttt{A}_{q}$ is the answer for query image and $\texttt{P}^{'''}_{q}$ is a new prompt designed to supply related information alongside the original question $\texttt{P}_{q}$.
Taking disease entity probing as example (in Figure~\ref{fig:tasks}(c)), $\texttt{P}^{'''}_{q}$ can be ``\texttt{Based on the query image, and the similar images and their reports: $(\texttt{img}_{q_1},  \texttt{A}_{q_1}, ..., \texttt{img}_{q_K}, \texttt{A}_{q_K})$, $\texttt{P}_{q}$},'' and $\texttt{P}_{q}$ is ``\texttt{Does the patient have [disease entity]?}''

\section{Experiment}

\subsection{Experimental Setups}

We selected RadFM~\cite{Wu2023TowardsGF}, an existing multi-image-trained Med-MLLM, as our base model to evaluate the effectiveness of V-RAG and our proposed fine-tuning tasks on multi-image-trained models. To assess the capability of making single-image-trained MLLMs V-RAG capable, we utilized LLaVA~\cite{Liu2023VisualIT} as the backbone model. We employed LoRA~\cite{Hu2021LoRALA} to fine-tune both LLaVA and RadFM on our designed tasks, applying a learning rate of 5e-5 for all fine-tuning tasks.

\subsection{Baselines}

We compare our method with the original Med-MLLM, RadFM, which does not include retrievals, with other baselines that do.
RAT~\cite{Sarto2024TowardsRA} and Img2Loc~\cite{Zhou2024Img2LocRI} are
identical methods which incorporate text associated with retrieved
similar images into the prompt.
RAR~\cite{Liu2024RARRA} also incorporates the text associated with retrieved
similar images, but it re-ranks those texts using the MLLM before generation.
We set $k = 5$ as the number of retrievals for every RAG-based method.

\begin{table*}[]
\begin{center}

\begin{tabular}{ccccccc}
\hline
\multicolumn{1}{c}{\multirow{2}{*}{\textbf{Method}}} & \multicolumn{3}{c}{\textbf{MIMIC-CXR}}  & \multicolumn{3}{c}{\textbf{MultiCaRe}} \\
\cline{2-7} 
\multicolumn{1}{c}{} & Precision & Recall & F1 & Precision &   Recall & F1 \\ \hline
				
RadFM &  \textbf{0.921}	& 0.206 &	0.381    & \textbf{0.972}       &   0.290    &  0.432   \\ \hline

+ RAR &   0.871	& 0.397 &	0.535   & 0.962 & 0.536 & 0.664 \\

+ RAT / Img2Loc&  0.760 &	\textbf{0.943} &	0.711  & 0.961 &	0.915	& 0.901   \\ \hline

 + V-RAG  &   0.770	& {0.920}	& 0.721 &      0.960 &	{0.952} &	{0.920} \\

+ V-RAG (fine-tuned)  &    0.790	& 0.921	& \textbf{0.751} &      0.961 &\textbf{0.999} &	\textbf{0.940} \\ \hline

\end{tabular}
\caption{Overall entity probing performance for different methods across two datasets. V-RAG's superiority shows the value of using complete retrieval information, both text and images. The improved performance of our fine-tuned V-RAG demonstrates enhanced image-text association abilities in Med-MLLM during V-RAG.}
\vspace{-5mm}
\label{tb:overall}
\end{center}
\end{table*}

\subsection{Datasets and Evaluation Metrics}

\subsubsection{Entity Probing}

We utilize two medical vision-language datasets: MIMIC-CXR~\cite{Johnson2019MIMICCXRAD}, containing chest X-ray images for radiology, and MultiCaRe~\cite{multicaredataset}, offering a variety of images across medical specialties.
We follow the official data split for MIMIC-CXR and randomly split MultiCaRe into train, validation, and test sets with a ratio of 8:1:1.
To construct VQA pairs for disease entity probing, we employ a biomedical named entity recognition (NER) model~\footnote{Stanza i2b2: https://stanfordnlp.github.io/stanza/biomed.html }~\cite{zhang2021biomedical} to extract all disease entities from the dataset's reports. 
We input these reports into LLMs (in our case, Llama-2 7B) to create closed-ended QA data with yes or no answers. 
For example, we ask ``\texttt{Does the patient have [disease entity] based on the report: [Report]?}", with answers formatted as \texttt{Yes/No}, simplifying error analysis. 
The use of LLMs allows for interpreting complex semantic structures within the text to accurately deduce potential answers. 
For instance, given the \texttt{[Report]}: ``An upper GI series on post-operative day 5 showing the duodenum ruling out stenosis.'' and \texttt{[disease entity]}: ``\texttt{stenosis}'', the LLM correctly answers ``\texttt{No.}''
By sampling segments from a medical report, we generate a sequence of concise, closed-ended questions paired with LLM-generated answers. The VQA dataset is then formed by associating these disease probing QA pairs with the original medical images.

For example, in MIMIC-CXR, we exclude entities in the ``INDICATION'' section of the report, as these reflect patient history or the reason for conducting the evaluation rather than X-ray findings.
Across both datasets, we found that less frequent entities are often already covered by more frequent ones (e.g., ``\texttt{right lower lobe atelectasis}'' as a particular kind of ``\texttt{atelectasis}''). Therefore, we map each entity to its shortest terminal subphrase occurring as an entity in the training set, to reduce redundancy and clarify entity frequency. 
For each test set of MIMIC-CXR and MultiCaRe, we parse 9,411 and 21,653 VQA pairs, respectively, with 385 and 10,434 distinct entities.
We use Precision, Recall, and F1 Score as the primary metrics to evaluate answer correctness in disease entity probing.

\subsubsection{Report generation}

We apply disease entity probing with V-RAG to mitigate hallucinations in generated text through a rewrite strategy. After a Med-MLLM generates an initial report of findings for an X-ray, the NER model extracts all disease entities from the generated report and from the reports of the $k$ most similar images.  For each entity, the query image is probed using the Med-MLLM with V-RAG.

The originally generated report and entity probing results are input to
a text-only LLM (Llama 3.1 70B chat), with the prompt:
\texttt{
Consider the following chest X-ray report from a junior radiologist:
-----begin report-----
[REPORT]
-----end report-----
A senior radiologist has inspected the X-ray image and answered the following questions:
-----begin questions----
[QUESTIONS AND ANSWERS]
-----end questions-----
Please rewrite the junior radiologist's report to reflect the senior radiologist's answers.}
We measure RadGraph-F1 scores \citep{delbrouck-etal-2024-radgraph}
of the findings of the original and revised reports.

\section{Evaluation Results}

\subsection{Overall performance for existing multi-image-trained Med-MLLMs}
We first evaluate V-RAG's performance for existing Med-MLLM that originally trained on multi-image datasets. Table~\ref{tb:overall} shows entity probing results comparing our method to baselines.
Across both datasets, V-RAG outperforms text-only RAG baselines in F1 scores. This improves the model's ability to extract relevant information for decision-making. Furthermore, with our proposed fine-tuning tasks, V-RAG (fine-tuned) achieves superior F1 scores over both baselines and the un-fine-tuned version. This shows that we have significantly enhanced Med-MLLM's capabilities by equipping it with robust image-text association skills.

\subsection{Ablation study}
We now conduct ablation studies to better understand our proposed method across various configurations.

\paragraph{Multimodal retrieval.} 
Table~\ref{tb:modality} shows the F1 scores of RAG (top-5) across different retrieval modalities. We observe that providing only similar images without text makes it challenging for Med-MLLM to extract entity information from visuals, though it offers marginal improvements over Med-MLLM without RAG. Adding text for similar images significantly enhances performance, highlighting the rich information provided by texts in entity probing.
By integrating both modalities, V-RAG effectively links retrieved texts and images, enabling more comprehensive decision-making and achieving the best performance. This underscores the importance of multimodal retrieval in V-RAG, rather than relying solely on text as most existing MLLM baselines.

\begin{table}[]
\begin{center}
\scalebox{0.8}{
\begin{tabular}{ccccc}
\hline
\multicolumn{2}{c}{{\textbf{Retrieval Modality}}} & \multicolumn{1}{c}{\multirow{2}{*}{\textbf{MIMIC-CXR}}} & \multicolumn{1}{c}{\multirow{2}{*}{\textbf{MultiCaRe}}} \\

\multicolumn{1}{c}{Image} & \multicolumn{1}{c}{Text}  & \\ 

\hline
 & &  0.381 & 0.432\\
\checkmark & &  0.705 & 0.735\\
& \checkmark & 0.711 & 0.901\\
\checkmark & \checkmark & \textbf{0.721} & \textbf{0.920}\\ \hline

\end{tabular}
}
\caption{Ablation study on RAG with different retrieval modalities. Improved F1 scores across both datasets shows the importance of integrating both image and text from retrievals to make informed decisions in V-RAG.}
\vspace{-4mm}
\label{tb:modality}
\end{center}
\end{table}

\paragraph{Fine-tuning tasks for V-RAG.}
To enhance V-RAG's performance, we proposed three fine-tuning tasks for Med-MLLM, each with 6,000 instances. In Table~\ref{tb:ablationfinetune}, we examine how different combinations of these tasks impact performance.
Initially, using only the $M_{vrag}$ dataset, we enable Med-MLLM to learn from extracted similar information, yielding performance gains that enhance the model's understanding of downstream V-RAG tasks.
Adding the image-text association tasks $M_{position}$ and $M_{focus}$ provides further gains, with $M_{position}$ offering more benefits due to the complexity of $M_{focus}$, which involves generating a full medical report and is more challenging to learn with limited data.

\begin{table}[]
\begin{center}
\scalebox{0.8}{
\begin{tabular}{cccccc}
\hline
\multicolumn{3}{c}{{\textbf{Fine-tuning Tasks}}} & \multicolumn{1}{c}{\multirow{2}{*}{\textbf{MIMIC-CXR}}} & \multicolumn{1}{c}{\multirow{2}{*}{\textbf{MultiCaRe}}} \\

\multicolumn{1}{c}{Position} & \multicolumn{1}{c}{Focus}
& \multicolumn{1}{c}{V-RAG}  & \\ 

\hline
 & & &  0.721 & 0.920 \\
 & &\checkmark  & 0.729 & 0.933 \\
 & \checkmark &\checkmark & {0.741} & {0.935} \\ 
\checkmark & &\checkmark & {0.748} & {0.937} \\ 
\checkmark & \checkmark &\checkmark & \textbf{0.751} & \textbf{0.940} \\ \hline


\end{tabular}
}
\caption{Ablation study of V-RAG using RadFM trained on various fine-tuning tasks. The F1 gains achieved through our three proposed tasks show improved image-text association abilities for existing multi-image-trained Med-MLLMs.}
\label{tb:ablationfinetune}
\vspace{-6mm}
\end{center}
\end{table}

\subsection{Analysis of entities across frequency levels}
\begin{figure}[t]
    \centering
    \setlength{\abovecaptionskip}{1mm}
    \includegraphics[width=0.9\linewidth]{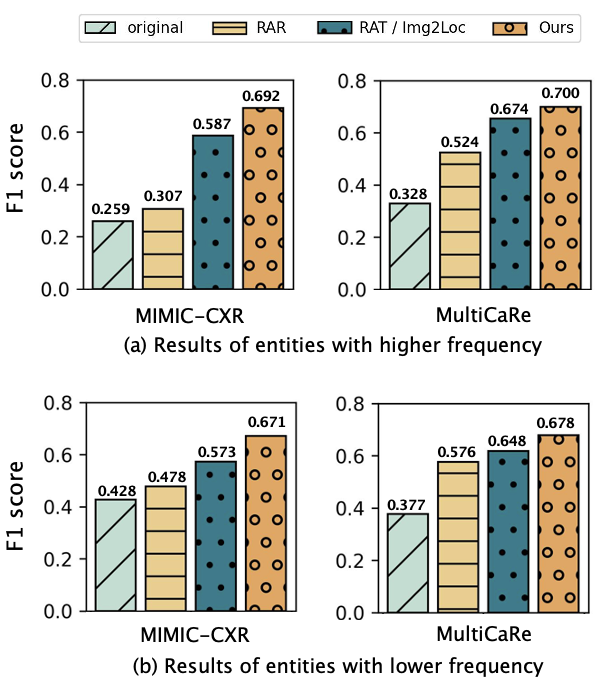}
    \caption{F1 performance of each method on disease entities with different frequencies. The superior performance of our method, particularly in probing rare entities, demonstrates its effectiveness and applicability in real-world scenarios.}
    \label{fig:freq}
\end{figure}

In addition to analyzing the overall entities in the test set, we conducted an analysis to see how they differ in appearance. We categorized the entities from the test set into the most frequent 50 and the less frequent ones, analyzing their performance separately. Rare entities were almost exclusively found in positive contexts, which created a label imbalance. To address this, we balanced the test sets for rare entities by adding additional negative probing questions for each entity until the number of positive examples equaled the number of negative examples. Negative examples were paired with a randomly chosen image, and we verified using Llama-2 that the associated report did not suggest the presence of the entity. We tested 1,000 samples for both frequent and rare entities across two datasets.
Figure~\ref{fig:freq} shows the F1 scores for each test set. Our V-RAG method outperforms both the original method and the RAG baselines in both settings. The improvement of V-RAG over other methods in the rare entity setting demonstrates the practical utility of our approach, emphasizing its effectiveness in utilizing information from multiple modalities to answer queries that neither the original model nor text-based RAG methods could address.

\begin{table}[]
\begin{center}
\scalebox{0.78}{
\begin{tabular}{cccc}
\hline
\multicolumn{1}{c}{\multirow{2}{*}{\textbf{Model}}} & \multicolumn{3}{c}{\textbf{MIMIC-CXR}} \\
\cline{2-4} 
\multicolumn{1}{c}{} & Precision & Recall & F1  \\ \hline
 LLaVA$_S$ &  \textbf{0.953}     &   0.475    &  0.604    \\ \hline
 LLaVA$_{\{M_{vrag}\}}$ & 0.914 &	0.867 &	0.852 \\
LLaVA$_{\{M_{focus} + M_{vrag}\}}$& 0.908 &	0.903 & 0.859 \\
LLaVA$_{\{M_{position} + M_{vrag}\}}$& 0.908	& {0.910} &	0.862\\
 LLaVA$_{\{M_{position} + M_{focus} + M_{vrag}\}}$ & 0.897 &	\textbf{0.944} &	\textbf{0.870}
\\\hline

\end{tabular}
}
\caption{Entity probing results for single-image-trained MLLM and MLLM enhanced with our proposed fine-tuning tasks. The superiority indicates that our tasks effectively make a MLLM V-RAG-capable.}
\vspace{-5mm}
\label{tb:llava}

\end{center}
\end{table}
\subsection{Can we make a single-image-trained MLLM V-RAG-capable?}
After observing the performance gains of V-RAG and our fine-tuning methods on multi-image pre-trained Med-MLLMs, we now explore whether single-image-trained MLLMs can also be enabled to perform V-RAG.
We extract all single image-text pairs from the MIMIC-CXR training set to create the single-image dataset $S$, resulting in 100,098 samples.
We then fine-tune LLaVA-v1.5-7B with Vicuna backbone~\cite{Liu2023ImprovedBW} on $S$ using LoRA for one epoch, resulting in a single-image Med-MLLM denoted as LLaVA$_S$.
From the single-image dataset $S$, we extract 10k samples for each fine-tuning task in Section~\ref{sec:v-rag-method}, creating the multi-image datasets $M_{position}$, $M_{focus}$, and $M_{vrag}$.
We then fine-tune LLaVA$_S$ on these tasks, producing the model LLaVA$_{\{task\}}$.

To evaluate our idea, we conducted entity probing on the MIMIC-CXR test set. For the single-image model LLaVA$_S$, we input a single test image to probe for a disease entity.
For the multi-image model LLaVA$_{\{task\}}$, we implemented V-RAG to assess its performance and determine if it can be effectively V-RAG capable with our designed tasks.
We set the context length of LLaVA to be 4096 and consider the top-3 retrievals for LLaVA$_{\{task\}}$ when performing V-RAG.

Table~\ref{tb:llava} shows the entity probing performance of single-image-trained MLLM and MLLM with multi-image capabilities resulting from our proposed fine-tuning tasks. 
For the single-image model LLaVA$_S$, we input a single test image and tasked the model with probing for a disease entity based on the given image. 
For the multi-image model LLaVA$_{\{task\}}$, we perform V-RAG to assess its performance. We set the context length of LLaVA to be 4096 and consider the top-3 retrievals for LLaVA$_{\{task\}}$ when performing V-RAG.
Results demonstrate that, with the support of our designed fine-tuning tasks, we enable the single-image-trained MLLM to effectively perform V-RAG.


\subsection{Improving generated reports}

\begin{table}[]
\begin{center}
\begin{tabular}{cccc}
\hline
\multicolumn{1}{c}{\multirow{2}{*}{\textbf{Reports}}} & \multicolumn{3}{c}{\textbf{RadGraph F1}} \\
\cline{2-4} 
\multicolumn{1}{c}{} & Simple & Partial & Complete  \\ 
\hline
Original & .163 & .145 & .102 \\
Revised & .194 & .172 & .118 \\
\hline
\end{tabular}
\caption{Revising reports with V-RAG entity probing results.}
\vspace{-5mm}
\label{tab:reportgen}
\end{center}
\end{table}

We have shown that disease entity probing provides a valuable clinical perspective on model outputs. 
However, since entity probing is typically not the final task for an MLLM, it is essential to demonstrate the utility of V-RAG in report generation.
We find that our strategy using Llama 3.1 70B Chat to rewrite the
generated reports using the V-RAG entity probing results yields 19\% relative
improvements in the simple and partial RadGraph-F1, compared to
the original findings, as shown in Table~\ref{tab:reportgen}.
These results highlight the practical benefits of V-RAG-enhanced entity probing, demonstrating its value not only in probing accuracy but also in improving the accuracy of generated medical reports.

\section{Conclusion}

When faced with a long report generation task, Medical Multimodal Large Language Models may exhibit biases and hallucinate details. We have introduced an entity probing method to examine these details, and shown that V-RAG improves entity probing accuracy for both frequent and rare entities. The lack of multi-image support in mainstream models has been a barrier to the adoption of V-RAG, leading almost all prior work to work only with the text corresponding to similar images.  Our special image-and-text fine-tuning tasks pave the way for multi-image-trained and single-image-trained models to become capable or more powerful at V-RAG, and we have shown that the use of both retrieved text and retrieved images benefits entity probing performance. Downstream, revision using entity probing with V-RAG can increase a report's accuracy on clinical details, improving the RadGraph-F1 score of a generated report.  Our research contributes towards more medically trustworthy MLLMs for healthcare applications.

\bibliography{aaai25}

\end{document}